\documentclass{bmvc2k}
\usepackage{subfigure}
\usepackage{amsfonts}

\usepackage[utf8]{inputenc}
\usepackage{paralist}
\usepackage{capt-of}
\usepackage{booktabs}
\usepackage{graphicx}

\makeatletter
\newcommand{\printfnsymbol}[1]{%
  \textsuperscript{\@fnsymbol{#1}}%
}
\makeatother
\begin{document}

\title{\emph{Texel-Att}: Representing and Classifying Element-based Textures by Attributes}

\addauthor{Marco Godi*}{marco.godi@univr.it}{1}
\addauthor{Christian Joppi*}{christian.joppi@univr.it}{1}
\addauthor{Andrea Giachetti}{andrea.giachetti@univr.it}{1}
\addauthor{Fabio Pellacini}{pellacini@di.uniroma1.it}{2}
\addauthor{Marco Cristani}{marco.cristani@univr.it}{1}

\addinstitution{
 Department of Computer Science\\
 University of Verona\\
 Strada le Grazie 15, Italy
}
\addinstitution{
 Sapienza, University of Rome\\
Rome,\\
 Italy
 {\textcolor{white}{.}} \\
{\textcolor{white}{.}} \\
{\textcolor{white}{.}} \\
{\textcolor{white}{.}} \\
{\footnotesize * These authors contributed equally to this work.} \\
}

\runninghead{Godi, Joppi}{Texel-Att: Representing [...] element-based textures}

\def\eg{\emph{e.g}\bmvaOneDot}
\def\Eg{\emph{E.g}\bmvaOneDot}
\def\etal{\emph{et al}\bmvaOneDot}


\maketitle

\begin{abstract}
Element-based textures are a kind of texture formed by nameable elements, the \emph{texels}~\cite{ahuja2007extracting},
distributed according to specific statistical distributions; it is of primary importance in many sectors, namely textile, fashion and interior design industry. State-of-the-art texture descriptors fail to properly characterize element-based texture, so we present \emph{Texel-Att} to fill this gap.  
Texel-Att is the first fine-grained, attribute-based representation and classification framework  for element-based textures. 
It first individuates texels, characterizing them with individual attributes; subsequently, texels are grouped and characterized through layout attributes, which give the Texel-Att representation.
Texels are detected by a Mask-RCNN, trained on a brand-new element-based texture dataset, \emph{ElBa}, containing 30K texture images with 3M fully-annotated texels. Examples of individual and layout attributes are exhibited to give a glimpse on the level of achievable graininess.
In the experiments, we present detection results to show that texels can be precisely individuated, even on textures ``in the wild''; to this sake, we individuate the element-based classes of the Describable Texture Dataset (DTD), where almost 900K texels have been manually annotated, leading to the \emph{Element-based DTD} (E-DTD).
Subsequently, classification and ranking results demonstrate the expressivity of Texel-Att on \emph{ElBa} and E-DTD, overcoming the alternative features and relative attributes, doubling the best performance in some cases; finally, we report interactive search results on \emph{ElBa} and E-DTD: with Texel-Att on the E-DTD dataset we are able to individuate within 10 iterations the desired texture in the 90\% of cases, against the 71\% obtained with a combination of the finest existing attributes so far. Dataset and code is available at \url{https://github.com/godimarcovr/Texel-Att}
\end{abstract}

\section{Introduction}
\label{sec:intro}
\begin{figure}[t!]
\centering
\subfigure[]{\includegraphics[width=0.25\textwidth,height=3cm]{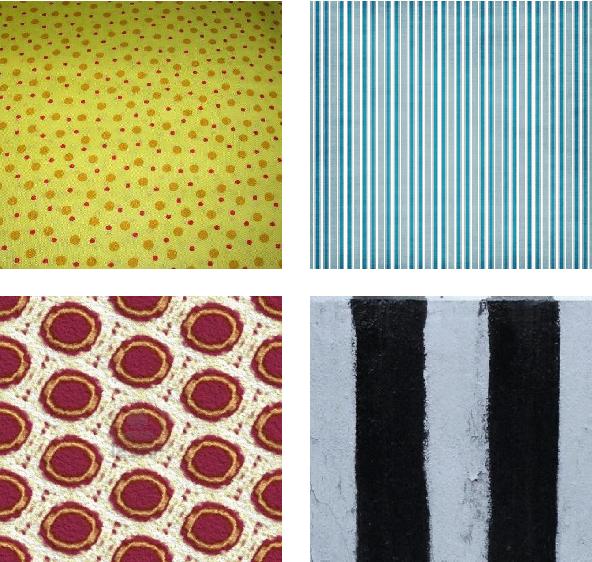}}
\hspace{0.2em}
\subfigure[]{\includegraphics[width=0.40\textwidth,height=3cm]{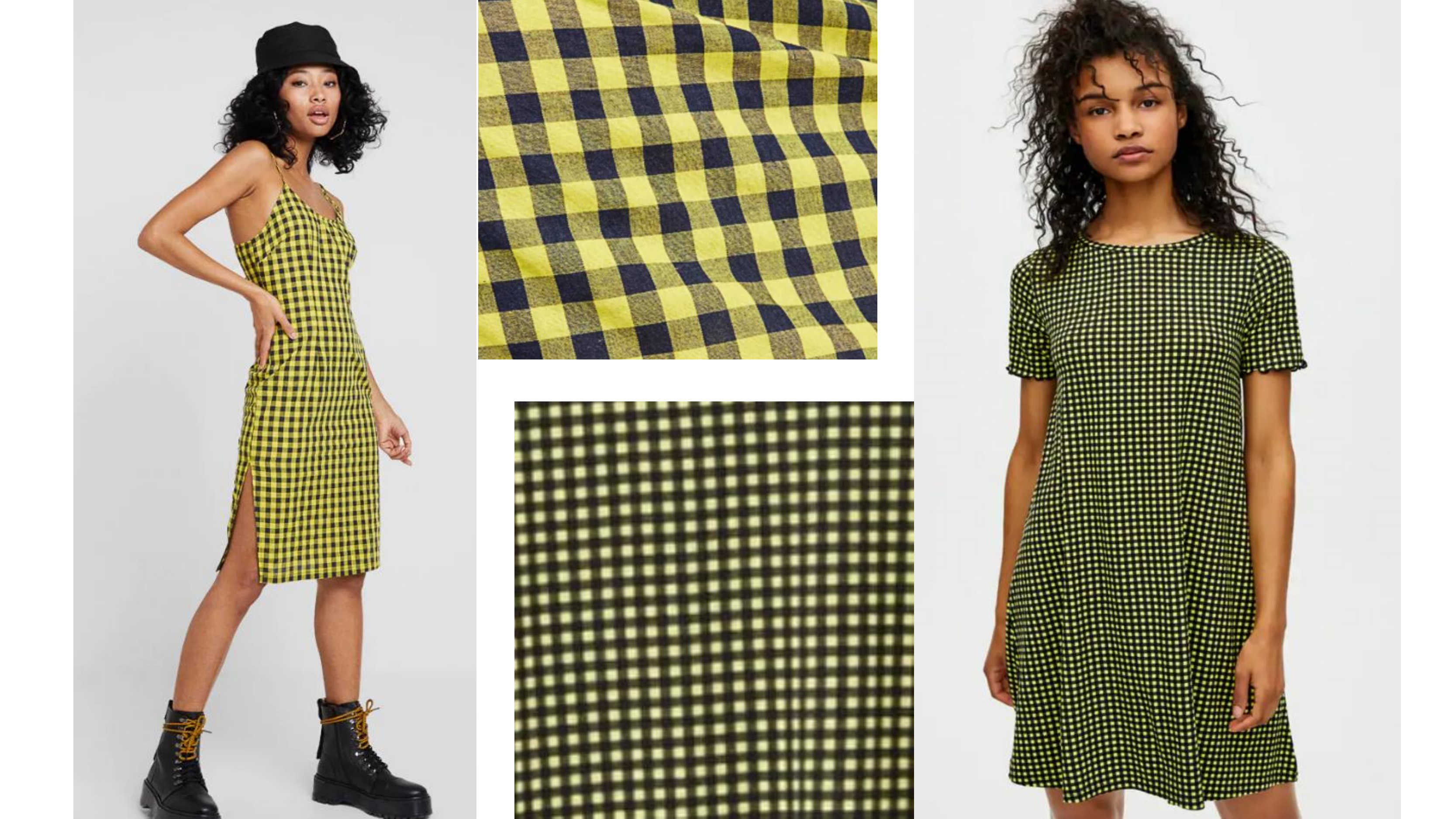}}
\hspace{0.2em}
\subfigure[]{\includegraphics[width=0.13\textwidth,height=3cm]{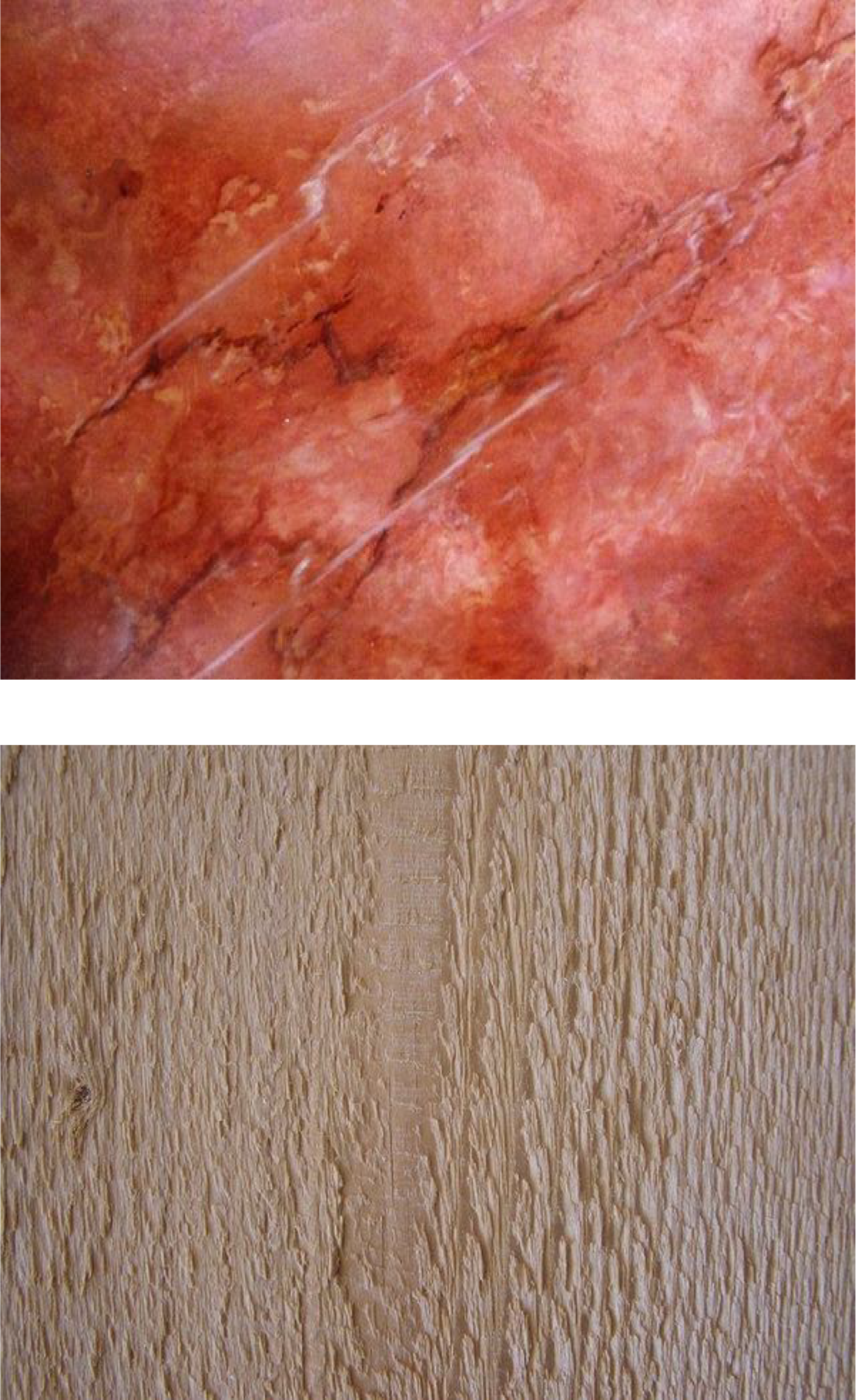}}
\hspace{0.2em}
\subfigure[]{\includegraphics[width=0.13\textwidth,height=3cm]{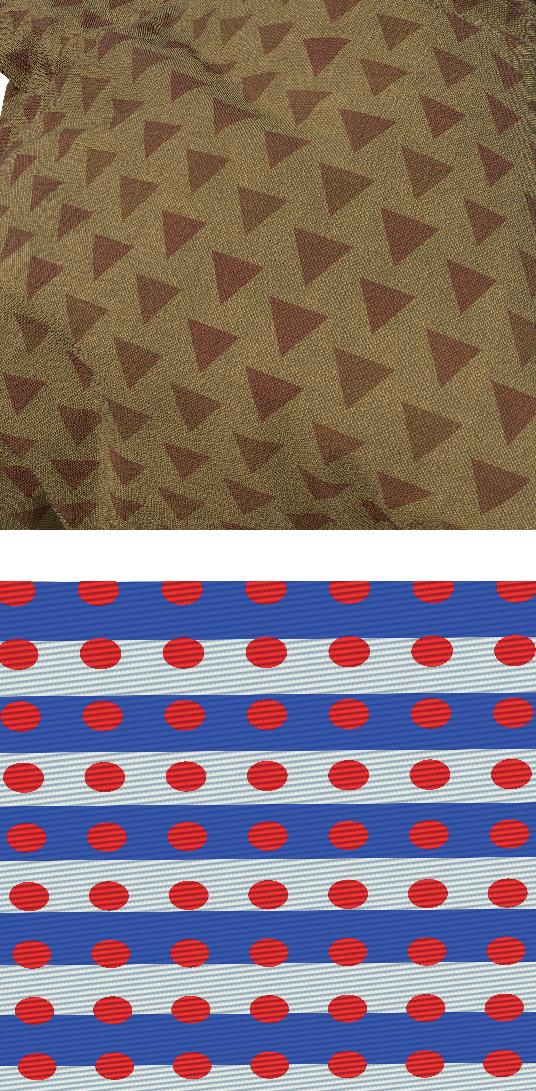}}
\caption{\small \label{fig:exte} (a) Examples of element-based textures in the DTD~\cite{cimpoi2014describing}: the \emph{dotted} (left) and \emph{banded} (right) classes are examples where texels are dots and bands, respectively; (b) Zalando shows for each clothing a particular on the texture; (c) examples of DTD~\cite{cimpoi2014describing} textures which are \emph{not} element-based: (\emph{marbled} on top and \emph{porous} on bottom); here is hard to find clearly nameable local entities; (d) examples of \emph{ElBa} textures: polygon on top, multi-class lined+circle texture on bottom.}
\end{figure}

\emph{Element-based} textures~\cite{ijiri2008example,ma2011discreet,ma2013dynamic,loi2017programmable} are 
textures formed by nameable recognizable elements, also dubbed \emph{texels}~\cite{ahuja2007extracting}, organized according to specific spatial distributions (see Fig.~\ref{fig:exte}a-b). 
They differ from those textures whose main characteristics are defined merely at a \emph{micro} scale, \emph{i.e.}, focusing on materials and material properties (see Fig.~\ref{fig:exte}c).

Element-based textures are of particular importance in the textile, fashion and interior design industry, with thousands of products (clothes, floor tiles, furniture) stored in vast catalogues or websites that the user (shopper, or designer) has to explore to find the preferred item to buy or to take inspiration from~\cite{kovashka2012whittlesearch,kovashka2015whittlesearch}; Fig.~\ref{fig:exte}b illustrates two examples taken from the popular e-commerce Zalando where, as many other fashion companies are doing now, clothing items are always accompanied by zoomed pictures highlighting textures.
\\
In these scenarios, describing textures and their compositional structure with communicative and intuitive features is of primary importance, in order to give a precise semantic content-based blueprint that allows a fast retrieval~\cite{smeulders2000content} or to design usable interfaces for professionals or casual users that create and print their own textile as in the web applications of \textsl{Patternizer, Contrado, Spoonflower, DesignYourFabric}, to quote some.\footnote{\url{https://patternizer.com/d0Wp}, \url{https://www.contrado.com/}, \url{https://www.spoonflower.com/} and \url{https://designyourfabric.ca/} respectively.}  

Attribute-based texture features~\cite{matthews2013enriching,roboticDatasetTex,cimpoi2014describing,surveytex2018} are explicitly suited to give textures nameable yet discriminative descriptions, and have been shown to excel in texture classification tasks.
The most-known texture attributes are the 47 perceptually-driven ones like \emph{dotted, woven, lined,} etc., that can  be learned on the Describable Texture Dataset (DTD)~\cite{cimpoi2014describing}.
\\
It is worth noting a limitation of these attributes: they describe the properties of a texture image \emph{as it was a whole atomic entity}: for example, in Fig.~\ref{fig:exte}a,
two different (element-based) attributes are considered, and specifically: \emph{dotted} (left) and \emph{banded} (right). For each attribute, two representative images are arranged in a column. Despite the same attribute that characterizes them, the images are strongly different: on the left, the top image has more, smaller dots; on the right, the top image has thin bands, while on the bottom bands are thick. 
In the Zalando examples (Fig.~\ref{fig:exte}b), both the clothes come with the same attribute (``checkered''), despite their diversity (squares have strongly different dimensions). 

In all cases, it is evident that for a finer expressivity one needs to focus on the recognizable \emph{texels} that constitute the textures.
\\
\\
In this paper, we present \emph{Texel-Att}, a fine-grained, attribute-based texture representation and classification framework for element-based textures. \\ 
The representation pipeline of Texel-Att 
detects first the single texels, assigning them \emph{individual attributes}. Subsequently, texels are grouped depending on the individual attributes, and groups of texels receive \emph{layout attributes}. 
 
Individual and layout attributes form the Texel-Att final description of the texture, that can be used for classification and retrieval. It is worth noting that the Texel-Att descriptor has no pre-defined dimensionality, as it depends on the attributes one does use. In this paper, we just give some examples to illustrate the general framework.

The texel detection is performed by a Mask-RCNN~\cite{he2017mask}, showing that element-based descriptions can be individuated with current state-of-the-art detection architectures (despite the fact that further improvements are foreseeable as we will discuss later). To train (and test) the detector, we design the first \emph{El}ement-\emph{Ba}sed texture dataset, \emph{ElBa}, taking inspiration by the many online catalogues and printing services cited above. \emph{ElBa} is composed of procedurally-generated realistic renderings, where we vary in a \emph{continuous} way element shapes and colors and their distribution, to generate 30K texture images with different local symmetry, stationarity, and density of (3M) localized texels, whose attributes are thus known by construction.
\\
In the experiments, we show that texels can be detected with high precision on \emph{ElBa}, and even on textures in the wild. To this sake, we consider the DTD~\cite{cimpoi2014describing}, selecting the 12 element-based classes and manually labeling the related almost 900K texels, originating the Element-based DTD (E-DTD).
Subsequently, we show texture classification results where
the pitfalls of current texture descriptors~\cite{cimpoi2016deep, tamura1978textural}do emerge, failing to classify toy examples where instead Texel-Att succeeds.  

Beyond the mere classification, Texel-Att descriptions induce reliable partial ordering so beating relative attributes~\cite{parikh2011relative} learned with alternative descriptors on ranking experiments. This fact is exploited in the final experiment to accelerate interactive image search~\cite{kovashka2015whittlesearch}, where a user is asked to find his desired texture among large repositories. Texel-Att allows to reach the goal with a lower number of steps, paving the way for commercial applications.

\section{The Texel-Att Framework}\label{sec:textiler}
\begin{figure}
\centering
   \includegraphics[width=\linewidth]{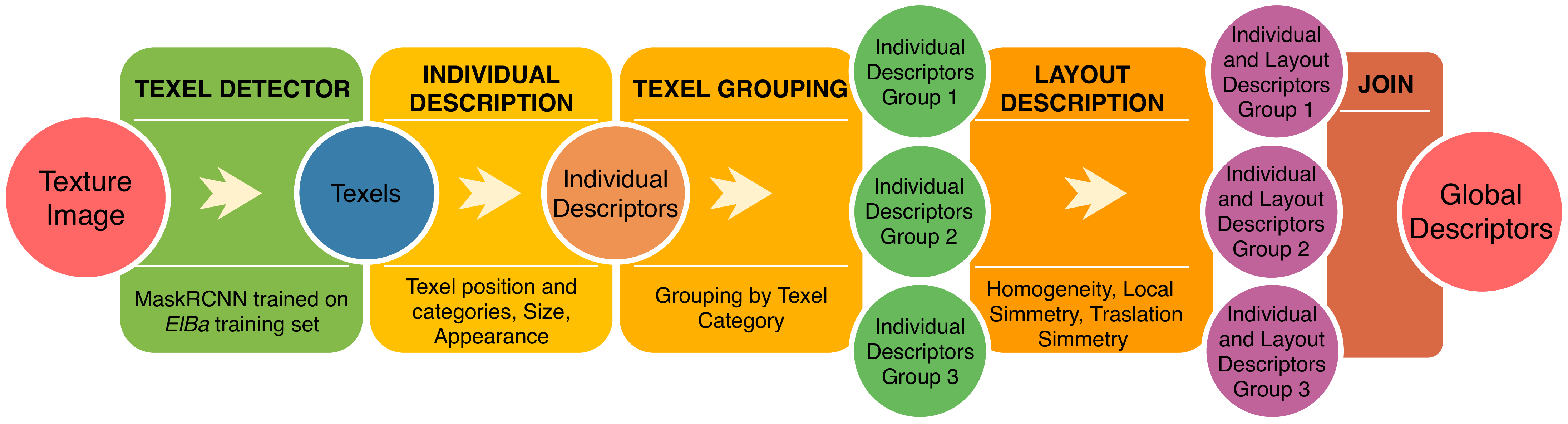} 
    \caption{\small \label{fig:scheme} Block diagram of the formation of the Texel-Att element-based texture descriptor. On the bottom of each plate, the specific choices made in this paper, which can be varied.}
\end{figure}

Fig. \ref{fig:scheme} shows a block diagram of the Texel-Att description creation pipeline. 
Briefly speaking, a customized region proposal method processes input images, extracting texels that subsequently are assigned with \emph{individual} attributes, \emph{i.e.},  
 labelled according to specific texel categories, and characterized with properties related to appearance and size. Individually labeled texels are then grouped, filtered (discarding non-repeated texels) and \emph{layout} attributes describing the spatial layout of groups are estimated. Individual and layout attributes form the composite Texel-Att descriptor.  
 In the following, each processing block is detailed.

\textbf{Texel Detector}. The texel detection is built on the Mask-RCNN~\cite{he2017mask} model, which localizes and classifies objects providing bounding boxes and  segmentation masks. 
We learn the model on the training partition of the \emph{ElBa} dataset, allowing to detect and classify as \emph{lines}, \emph{circles}, \emph{polygons} potentially each texel in a given image (see Sec.~\ref{sec:ElBa_dataset}). The message here is that the texels, whose detection a few years ago was quite complicated and limited to specific scenarios (\emph{i.e.}, lattices~\cite{gui2011texel,liu2015patchmatch}), are now easily addressable in whatsoever displacement.  

\textbf{Individual description of texels.} Each detected texel is characterized with attributes related to shape properties and human perception, and in particular: (i) the \textit{label} provided by the Mask-RCNN, indicating its shape; (ii) main \textit{color} given by a color naming procedure~\cite{van2009learning} (with 11 possible colors); (iii) element \textit{orientation}, if any; (iv) element \textit{size}, estimated as the area of the region mask.
Textures can be characterized by statistics computed on these features (averages or histograms, see in the following). It is worth specifying that different individual features and statistics could be adopted; in fact here we are not looking for ``the best'' set of features, but we are showing the portability and effectiveness of the general framework.

\textbf{Texel Grouping}
The goal is to cluster texels with the same appearance, to capture choral spatial characteristics via layout attributes. Here we simply group texels according to the assigned class labels (\emph{circle}, \emph{line} or \emph{polygon}).
Only groups including at least 10 texels are kept, the other detections are removed. 

\textbf{Layout description of texels.}
To describe spatial patterns of each texel group, we measured attributes
related to the spatial distribution of the texels' centroids.
Among the huge literature in statistics on spatial points patterns' analysis 
to evaluate 
 randomness, symmetry, regularity and more~\cite{diggle1983statistical,velazquez2016evaluation,baddeley2015spatial}, we selected a simple yet general set of measurements. They are:
(i) point \textit{density}, \emph{e.g.} the number of texels per area unit (for circles and polygons) or line density,
\emph{e.g.} the number of lines/bands along the direction perpendicular to their principal orientation (for lines).
(ii)  Quadratic counts-based \textit{homogeneity} evaluation~\cite{illian2008statistical}: it amounts to divide the original image into 100 patches and perform a $\chi^2$ test to evaluate the hypothesis of average point density in the sub-parts. Also in this case for lines, we estimated a similar 1D feature on the projection.
(iii) Point pair statistics~\cite{zhao2011translation}: we estimate the point pair vectors for all the texel centers and then use them to estimate the histogram of \textit{vectors orientation}.
(iv)  \textit{Local symmetry}: for circles and polygons we considered the centroids' grid and measured average reflective
self-similarity of 4-points neighborhoods of points after their reflection around the central point. The distance function used is the average point distance, normalized by neighborhood size.
Similarly, we measured \textit{translational symmetry} by considering 4-point neighborhoods of the centroids and measuring the average minimum distance of their points from closest points in the grid after translations of the vectors defined by point pairs in the neighborhood. Symmetry descriptors are computed on 1D projections for line texels.
The complete pattern descriptor is finally built joining texel attributes, spatial pattern attributes and the color attributes of the \textit{background}.
The dimensionality for each of these attributes is reported in Tab.~\ref{tab:descriptorSize}. 1-dimensional attributes are averages of all the extracted values, while multi-dimensional ones are histograms. More details on these can be found as supplementary material. 
The Texel-Att descriptor is composed by concatenating the attributes and Z-normalizing each one of them.
\begin{table}[t]
\small
\begin{flushleft}
\begin{minipage}{.35\textwidth}
\scalebox{0.6}{
\begin{tabular}{ccccc}
\toprule
Label & Color & Orientation & Size & \textbf{Total}\\
Histogram &  & Histogram &  & \\
\toprule
3 & 11 & 3 & 1 & 18\\
\bottomrule
\end{tabular}
}
\end{minipage}
\begin{minipage}{.60\textwidth}
\scalebox{0.67}{
\begin{tabular}{ccccccc}
\toprule
Density & Homogeneity & Vector  & Local  & Traslational  & Background  & \textbf{Total}\\
 &  &  Orientations &  Symmetry &  Symmetry &  Color & \\
\toprule
1 & 1 & 3 & 1 & 1 & 11 & 18\\
\bottomrule
\end{tabular}
}
\end{minipage}

\end{flushleft}
\small
\caption{\small \label{tab:descriptorSize} Dimensionality of descriptor attributes. On the left, the attributes computed from the individual characterization of texels; on the right, attributes computed from statistics resulting from the spatial layout. The total dimensionality of the descriptor is 36. }
\end{table}

\begin{figure}[h!]
\centering
\includegraphics[width=0.9\linewidth]{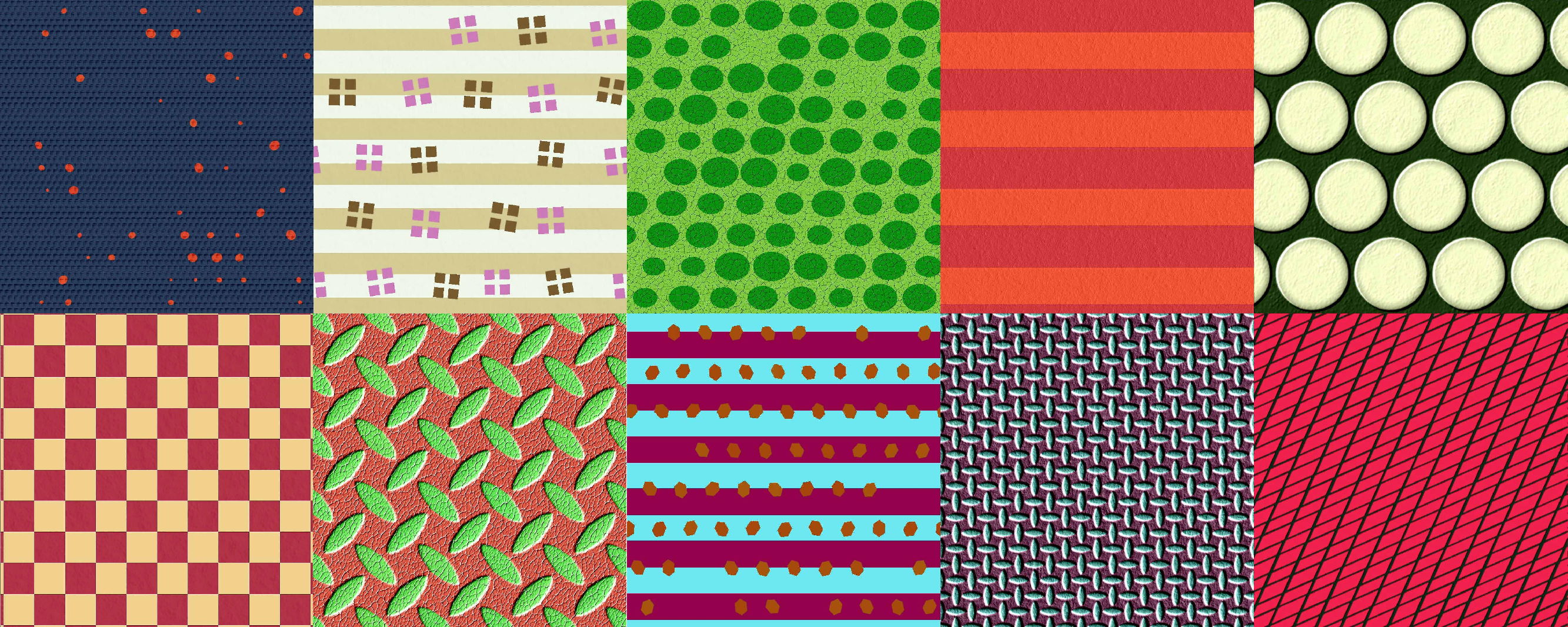}
\caption{\small \label{fig:toy_images} Images from the \emph{ElBa} dataset.}
\end{figure}

\section{Element-Based Texture Datasets: \emph{ElBa} and E-DTD}
\label{sec:ElBa_dataset}
While element-based textures are common 
and relevant to many practical applications (see Fig.~\ref{fig:exte}), no public database focused on this texture domain is available. Existing databases as the DTD~\cite{cimpoi2014describing} include some examples of these textures (Fig.~\ref{fig:exte}(a)), but these are mixed with other texture types. 
For these reasons, we build \emph{ElBa}, the first dataset of element-based textures, and \emph{E-DTD}, the element-based portion of the DTD dataset.

\emph{ElBa} includes synthetic photo-realistic images, like those shown in Fig.~\ref{fig:exte}(d).
The advantages of dealing with synthetic textures 
are the precise annotations for texels by construction,  and the possibility to train adequately a deep classifier, since training with synthetic data is a common practice~\cite{tremblay2018training,barbosa2018looking}. 
In particular, we propose a parametric synthesis model where we vary both texel \emph{individual} (addressing the single texel) and \emph{layout} attributes (describing how groups of texels are mutually displaced). 
As for individual attributes, we vary texel shape, size and orientation and color as follows.
For the shape, inspired by the 2D shape ontology of~\cite{niknam2011modeling},
we consider general regular entities as \emph{circles}, \emph{lines}, \emph{polygons} (squares, triangles, rectangles)
since they can be thought of as approximations of more complicated shapes and because they encompass a large variety of geometric textiles.
Size and orientation are varied linearly over a bounded domain. Colors are chosen from harmonized color palettes to better represent realistic use of colors.
Texels are placed in 2D space based on a variety of layouts that can be described succinctly using symmetries. We consider both linear and grid-based layouts where the layout attributes are defined by one or two non-orthonormal vectors that define the translation between texels in the plane. This simple description represents several tiling of the plane and their corresponding patterns. We consider both regular and randomized distributions, where the randomization is performed by jittering the regular grid. 
By using jittering, we create a continuum between regular and non-regular distributions, and by varying the jitters per-texel we can change the stationarity.

Importantly, we take into account distributions of more than one element type 
arbitrarily combined in the plane. For example, we can create dotted+striped patterns. Each texel type has its own spatial layout attributes effectively creating arbitrary multi-class element textures (Fig.~\ref{fig:exte}(d), Fig.~\ref{fig:detRes}-right).

We generate the images of \emph{ElBa} using state-of-the-art computer graphics tools. We use Substance Designer for pattern generation
\footnote{\url{https://www.allegorithmic.com/}}. Substance gives high-quality pattern synthesis, easy control and high-quality output including pattern antialiasing. High frequency patterns simulating realistic materials are added to the generated images.

This procedure led to a rough total of 30K of diverse textures rendered at a resolution of $1024 \times 1024$. The texture design process automatically provides ground-truth data for our analysis including texels masks, texels bounding boxes, and spatial distribution attributes. In total this amounts to around 3M annotated texels. 
It is very important to note that, differently from the other datasets used in the texture analysis domain, \emph{ElBa} does not come with a rigid partition into classes: semantic labels that define relevant classification tasks like those used in our tests (Sec.~\ref{sec:exp:classification}) can be derived by texels' attributes or by experiments with subjects.

The complete dataset is split in a training part (90\%), used to train the network model and a test part (10\%), used to validate the classification and recovery experiments.

\begin{figure}
\centering
   \includegraphics[width=0.9\linewidth]{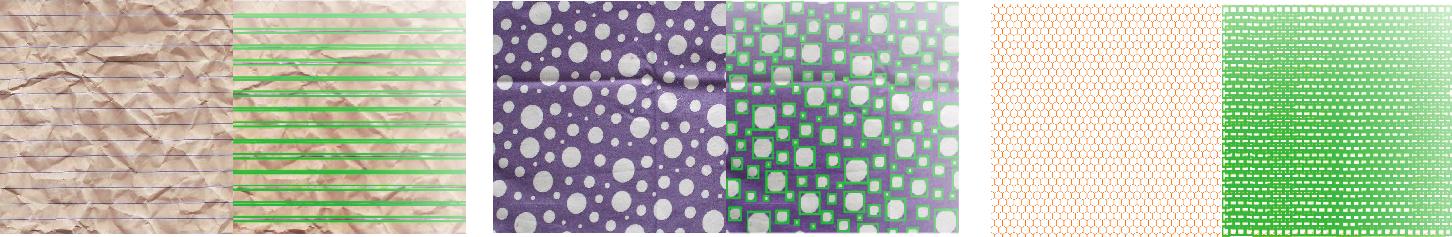} 
    \caption{\small Examples of E-DTD $\,$ annotations: ground truth green bounding boxes overlayed to images of the classes \emph{lined, dotted, honeycombed}, respectively. \label{fig:texture_labeling} }
\end{figure}

To demonstrate our approach on real images, we created the Element-based DTD (E-DTD) as follows. From the DTD, we extracted textures that are element-based, i.e. characterized by a distribution or recognizable repeated texels with limited perspective distortions. 
We manually annotated the bounding boxes of each texel. 
E-DTD includes 1440 images belonging to 12 of the original DTD classes: \emph{Banded, Chequered, Dotted, Grid, Honeycombed, Lined, Meshed, Perforated, Polka-dotted, Studded, Waffled. }
These classes have been selected by 7 experts (3 graphic designers and  2 fashion experts and 2 computer scientists) with all of them agreeing on their inclusions. DTD classes with partial consensus have not been inserted into E-DTD. 
The annotation of texels was carried out by Mechanical Turk, borrowed from the three-phase ImageNet crowdsourcing annotation protocol~\cite{su2012crowdsourcing}. The protocol consists of (1) a drawing phase, 
(2) a quality verification phase, where a second worker validates the goodness of the bounding boxes and (3) a coverage verification phase where a third worker verifies whether all object instances have bounding boxes. 
The annotation process produced around 900K texels annotations, some of which are shown in Fig.~\ref{fig:texture_labeling}.
It is important to note the very 
diverse types of bounding boxes, from very long and thin (addressing line texels) $745\times 5$ pixels bounding boxes to very small $5\times 5$ pixels bounding boxes (on tiny circles). 

\section{Experiments}\label{sec:exp}
Experiments focus on four aspects: 1) \emph{detection of texels}, where we show that finding texels is nowadays possible, with a Mask-RCNN trained on \emph{ElBa}; 
2) \emph{classification}, where we point out the failure of state-of-the-art descriptors in distinguishing textures which are clearly diverse against our Texel-Att that instead is succeeding;
3) \emph{ranking}, where we demonstrate that Texel-Att representation ranks texture w.r.t. expressive yet fine-grained attributes; 
4) \emph{image search} where the Texel-Att attributes are exploited for accelerate human-in-the-loop image search~\cite{kovashka2012whittlesearch} onto large image corpora. 
    
\subsection{Detection of texels}\label{sec:exp:detection}
\begin{figure}[ht]
\begin{minipage}[b]{.61\textwidth}
\centering
\includegraphics[width=0.48\textwidth]{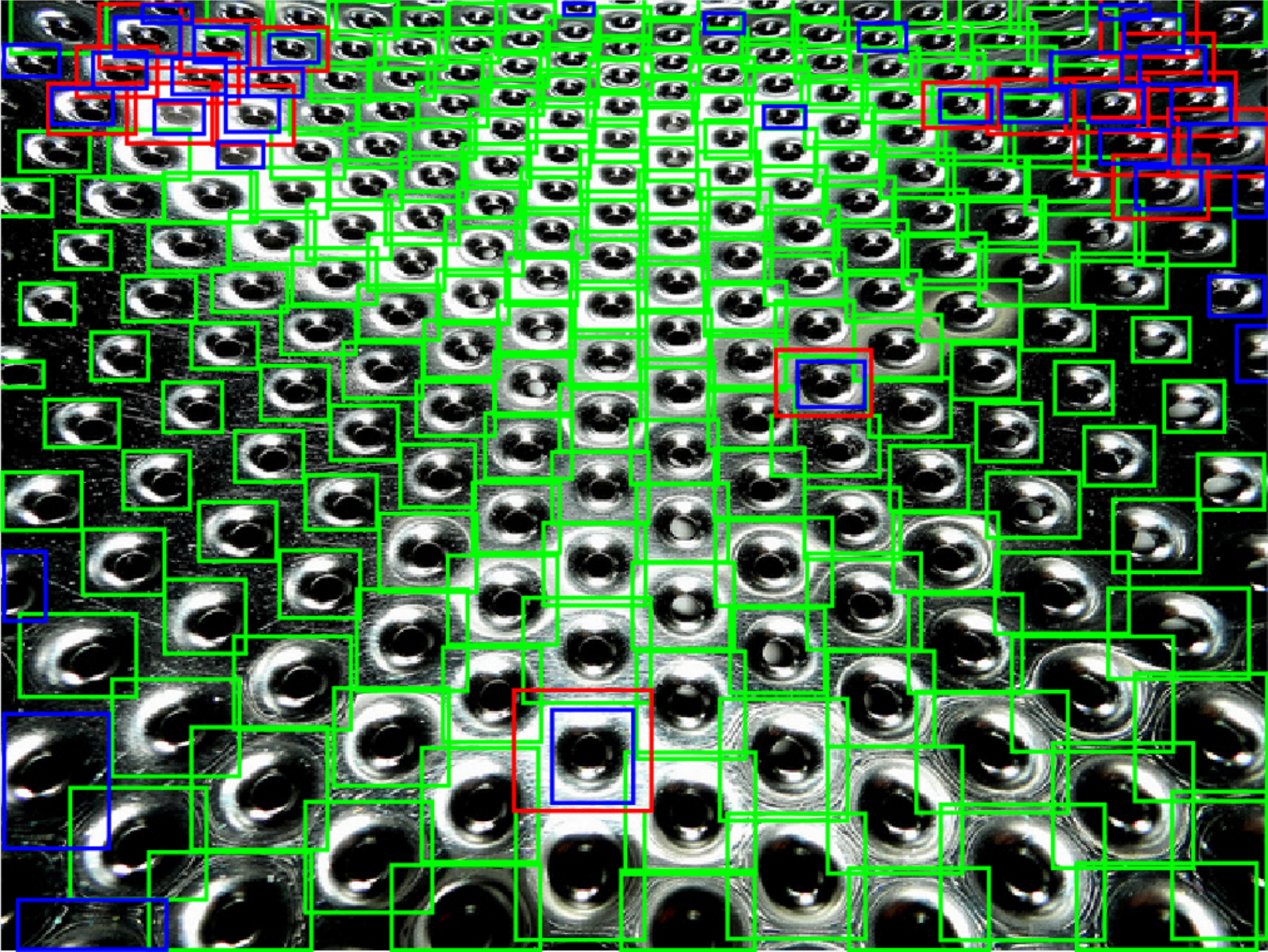}
\includegraphics[width=0.48\textwidth]{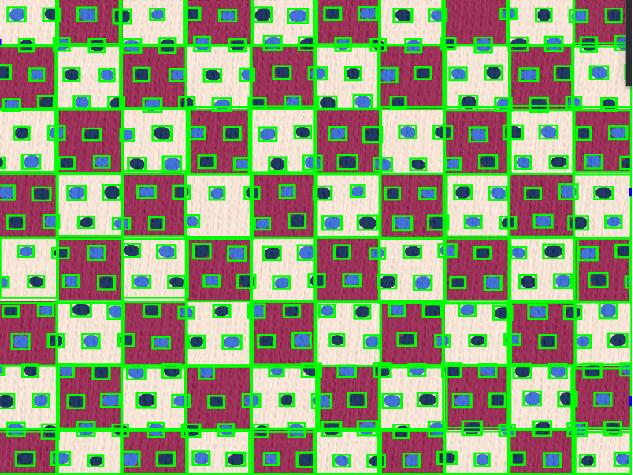}
\caption{\small \label{fig:detRes} Texel-Att detection qualitative results on both E-DTD (left) and \emph{ElBa} (right) datasets. In green the correct detections, in red the false positives (19 in the first, 0 in the second) and in blue the false negatives (35 in the first, 3 in the second). The AP(IoU=.50) are 0.81 and 0.99, respectively.}
\end{minipage}
\hfill
\begin{minipage}[b]{.37\textwidth}
\centering
\includegraphics[width=0.9\textwidth]{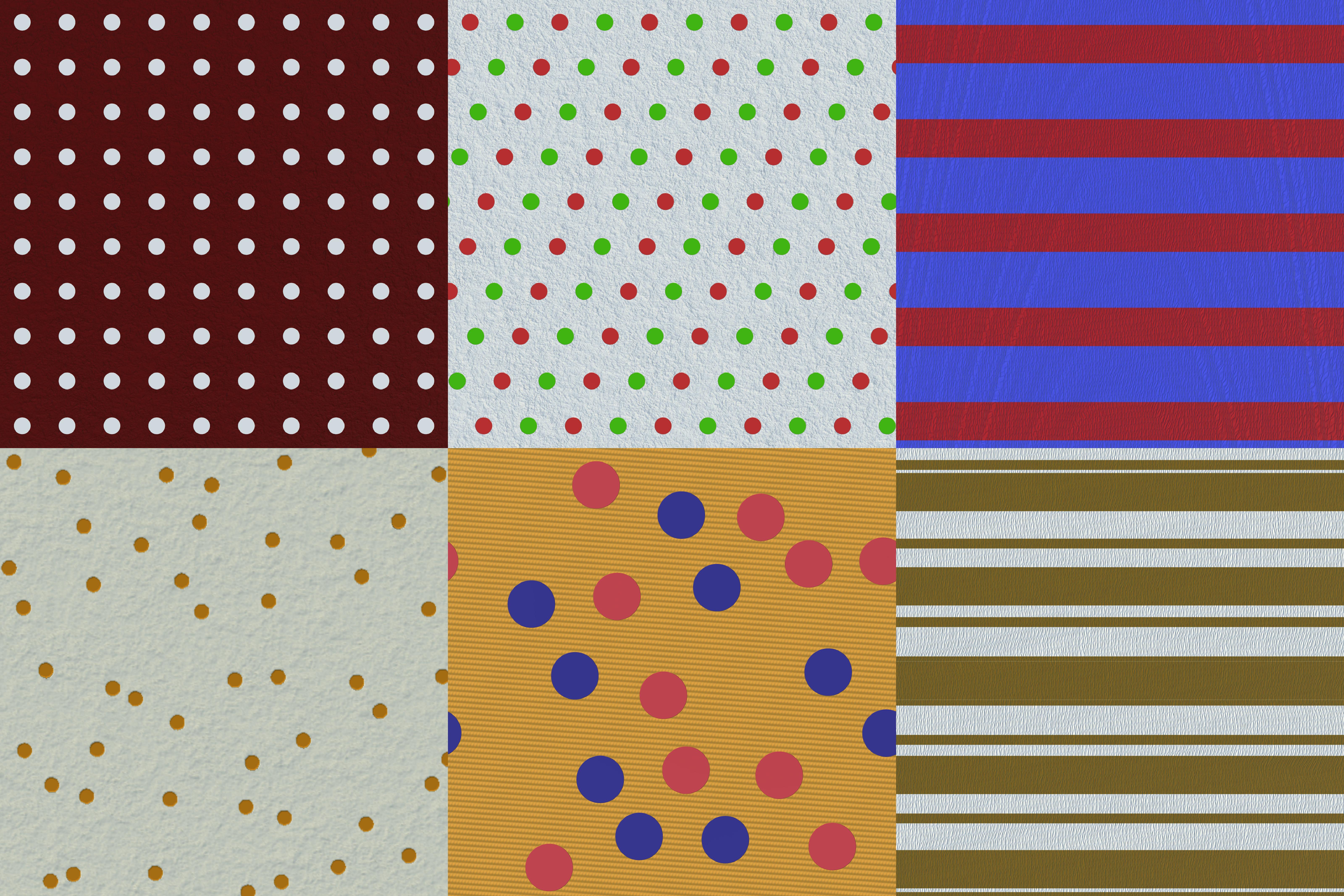} 
\caption{\small \label{fig:toy_images} Images from the \emph{ElBa} dataset, in columns: mono-colored regular and random circles, bi-colored regular and random circles,  uniform and non-uniform lines.}
\end{minipage}
\end{figure}

Detection performances have been computed on the testing partition of \emph{ElBa} and on the whole E-DTD. The Mask-RCNN model used in these experiments has been trained on the training partition of \emph{ElBa}.
Fig.~\ref{fig:detRes} reports some Texel-Att detection qualitative results, while  Tab.~\ref{tab:detectionResults} reports  \emph{per-image} average precision (AP): in practice, AP is computed \emph{for each image}, and averaged over all the images, since we are interested in capturing how much \emph{all of the texels of a single image} are detected, since it is crucial for computing the Texel-Att attributes afterwards. E-DTD dataset gives lower results, since it contains images with dramatic perspective deformation (see Fig.~\ref{fig:detRes}(a)), which was not a factor in the \emph{ElBa} training data. Despite this, the next experiments show that such detection performance is enough to estimate attributes with high accuracy.  Mask-RCNN trained on COCO gives dramatically low results (mAP = 1.75e-6), due to the completely different scenario, not reported in Tab.~\ref{tab:detectionResults} for clarity. One may ask how Texel-Att detection works on a texture which is not element-based, like Fig.~\ref{fig:exte}d. Few tests showed that the confidence of the detections, in that case, is definitely lower than in the element-based case, omitted here for the lack of space.  
\begin{table}[thb]
\begin{minipage}{.35\textwidth}
\begin{center}
\scalebox{0.85}{
\begin{tabular}{lccc}
\toprule
\textbf{Dataset} & mAP & AP50 & AP75\\
\midrule
E-DTD & 0.53  & 0.63   & 0.40\\
\emph{ElBa} &  0.91   &0.92 &0.90\\ \bottomrule
\end{tabular}}
\end{center}

\captionof{table}{\small \label{tab:detectionResults} Detection \emph{per-image} average precision (see text) on E-DTD and \emph{ElBa} datasets.}
\end{minipage}
\hspace{1.5em}
\begin{minipage}{.60\textwidth}
\begin{center}
\scalebox{0.85}{
\begin{tabular}{lccc}
\toprule
\textbf{Classes} & Tamura~\cite{tamura1978textural} & FV-CNN~\cite{cimpoi2016deep} & Texel-Att\\
\midrule
Line uniformity & 70.60  & 66.80   &\textbf{85.30}\\
Circle positioning &  54.86 & 53.01 & \textbf{97.33}\\ 
Circle coloring &  50.19 &  52.94 & \textbf{93.42}\\ \bottomrule
\end{tabular}
}
\end{center}
\caption{\small \label{tab:classificationResults} Classification accuracy in three different binary tasks with three different approaches.}

\end{minipage}
\end{table}
\vspace{-1em}

\subsection{Classification}\label{sec:exp:classification}
On texel classification into \emph{circles, polygons, lines} categories, Mask-RCNN scores a 99.85\% of accuracy on almost 550K of correctly detected (IoU>0.5) texels. Class labels are used to organize texels into groups, as described in Sec.~\ref{sec:textiler}. Groups are described with layout attributes, and this completes the Texel-Att description.
Three simple experiments on binary classification demonstrate the expressivity of Texel-Att, each one focusing on 200 \emph{ElBa} images 
having strongly different attributes, that is \emph{single-color VS bi-color circles}, \emph{regularly VS randomly positioned circles} and  \emph{lines with uniform or non-uniform width} (see Fig.~\ref{fig:toy_images}).
Cross-validated 5-fold experiments
compare the 36-dimensional Texel-Att description (see Tab. \ref{tab:descriptorSize}) against
CNN+FV (65536-dimensional)
and the Tamura~\cite{tamura1978textural} classic texture description.
All of the three descriptors are fed into linear SVMs. Accuracies are shown in Tab. \ref{tab:classificationResults}.
Texel-Att obtains the best results as it 
captures higher visual semantics, i.e., the texels and how they are mutually related. 
FVs and Tamura features are not able to capture objects and spatial layout, focusing on filter outputs or directly on pixel values.

On E-DTD, Texel-Att description individuates strongly different textures within the same class,
ideally defining further, finer-grained level of classes it can separate. For example the \emph{dotted} and \emph{banded} classes (see Fig.~\ref{fig:exte}) are now further specified and classified considering big or small dots, regular or irregular bands. For lack of space, we prefer to skip the fine-grain classification experiments on the E-DTD (where we systematically beat the other descriptors) and instead present in the next section the more compelling experiments on ranking.    
\subsection{Ranking}\label{sec:exp:rel_attributes}
Other than capturing fundamental properties of textures (for example, having regularly VS randomly placed texels, see the previous section), Texel-Att attribute values can be used to rank textures. Attributes that can be ranked are the basis for human-in-the-loop search strategies~\cite{kovashka2015whittlesearch}, so it is crucial that the ranking is reliable. Ideally, with the ground-truth texel detection the ranking via Texel-Att attributes will be perfect. In this experiment we evaluate how our detection step corrupts the ranking, and whether the ranking can be better estimated with learning based strategies, avoiding the detection step.  

For simplicity, we consider here partial ranking; an attribute that induces partial ranking is said \emph{relative}~\cite{parikh2011relative}; formally, given a set of images I=$\{i,j\}$ and an \emph{ideal} Texel-Att attribute $a$ (\emph{i.e.}, computed on the ground-truth texel detection) , there exists a partial order relation
r$_a^*$ such that $i > j \iff r_a^*(i) > r_a^*(j)  \land  |r_a^*(i) - r_a^*(j)| > \gamma_a$. 

The goal of this experiment is to estimate a function r$_a$(i) as close as possible to r$_a^*$.
Following the protocol of~\cite{parikh2011relative} the \emph{ranking accuracy} of the function r$_a$ is defined as the percentage of pairs correctly ordered by the r$_a$ function over all the possible pairs in the set of images.

Two are the strategies we compare to estimate r$_a$:  the first is the Texel-Att pipeline, which measures the attribute on top of the texel detections. The second is 
the relative attribute estimation of~\cite{parikh2011relative} using the FV-CNN~\cite{Cimpoi_2015_CVPR} descriptor as input. For this second strategy we can assume x$_i$ as the feature vector in $\mathbb{R}^n$ for the image $i$. In this case r$_a$ is \emph{estimated} by a ranking SVM, following the guidelines in~\cite{parikh2011relative}. We assume r$_a = w^{T}_ax_i$ so that the output of the modeling is the unknown vector $w$. The model is trained using the set of ordered pairs of images $O_a = \{(i,j)\}$ where $(i,j) \in O_a \Rightarrow	i \succ j$ and the set of un-ordered pairs of images $S_a = \{(i,j)\}$ where $(i,j) \in S_a \Rightarrow	i \sim j$.

We perform this experiment on the \emph{ElBa} dataset (using the partitioning defined in Sec.~\ref{sec:ElBa_dataset}) and the E-DTD dataset (randomly choosing 90\% of the images as training set and the rest as testing set). We consider one attribute at a time. Ground truth r$_a^*$ (i.e. ordered and un-ordered pairs) are computed from the ground truth detections also used in Section~\ref{sec:exp:detection}.
The ranking accuracy across all attributes is shown in Tab.~\ref{tab:resRelativeAttr}. It can be clearly seen that computing explicitly texel detection (\emph{i.e.}, following the Texel-Att pipeline) is the best strategy to rank textures according to the proposed attributes. For space reasons, only three attributes are shown on the table. The remaining ones are reported in the supplementary material.

\begin{table}[t]
\small
\begin{center}
\scalebox{0.85}{
\begin{tabular}{lcc}
\toprule
\textbf{Attributes} & Rank SVM~\cite{parikh2011relative} & Texel-Att\\
\midrule
Area    &65.52   &\textbf{85.71}\\
Local Symmetry    &54.72   &\textbf{64.29}\\
Homogeneity  &79.91   &\textbf{90.29}\\ 
\toprule
Mean    &67.56   &\textbf{81.68}\\
\bottomrule
\end{tabular}}
\hspace{0.4cm}
\scalebox{0.85}{
\begin{tabular}{lcc}

\toprule
\textbf{Attributes} & Rank SVM~\cite{parikh2011relative} & Texel-Att\\
\midrule
Area    &70.16   &\textbf{93.86}\\
Local Symmetry    &71.06   &\textbf{81.71}\\
Homogeneity  &77.06   &\textbf{96.86}\\ 
\toprule
Mean   &65.95   &\textbf{84.55}\\
\bottomrule
\end{tabular}}
\end{center}
\small
\caption{\small \label{tab:resRelativeAttr} Ranking accuracy of relative attributes on E-DTD (left) and \emph{ElBa} (right) datasets.}
\end{table}

\subsection{Texture Interactive Search}\label{sec:exp:wsearch}
\begin{figure}
    \includegraphics[width=1.0\linewidth]{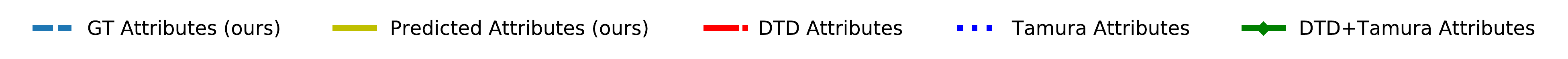}
   \includegraphics[width=0.24\linewidth, height=0.18\linewidth]{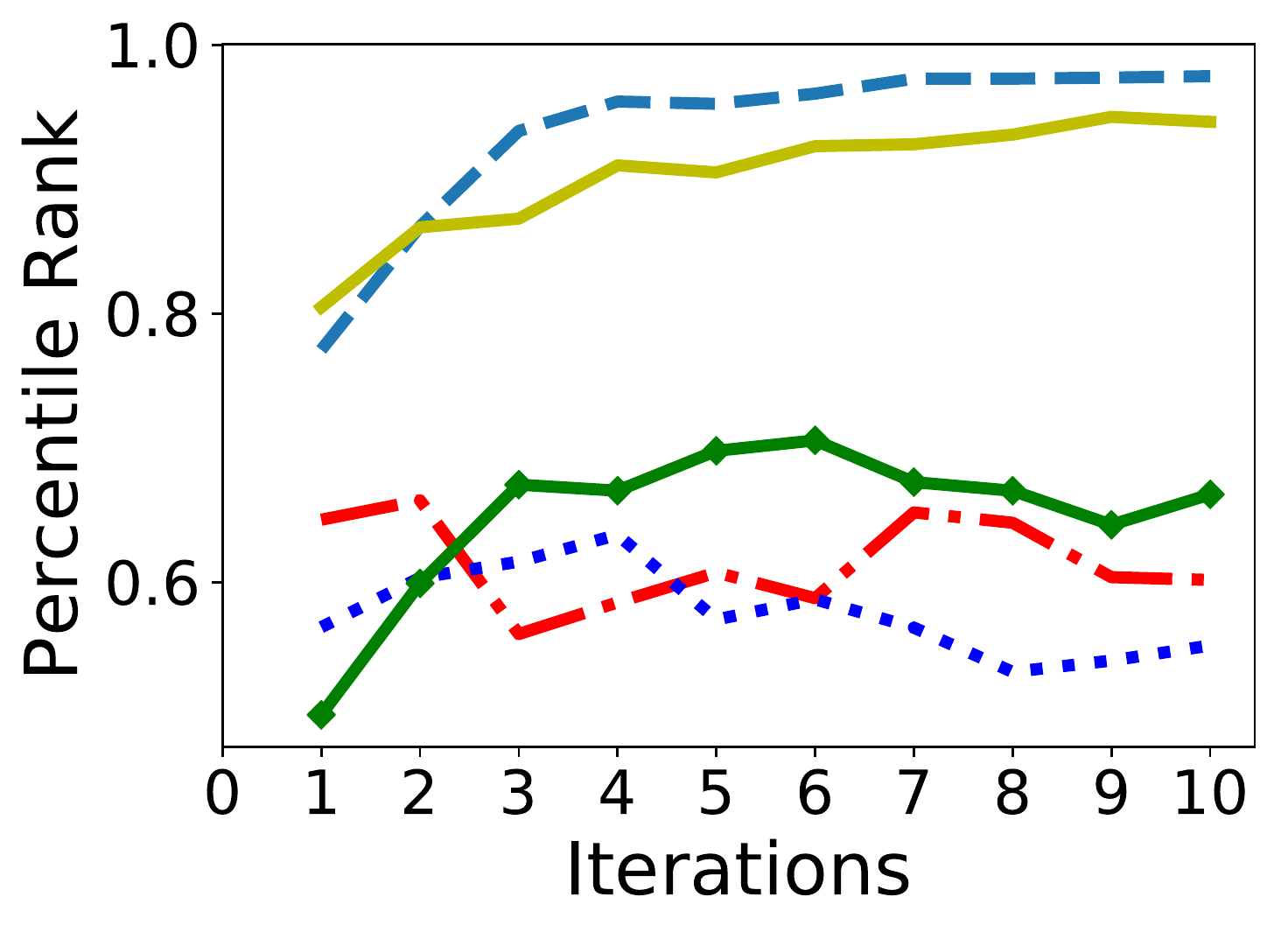}
   \includegraphics[width=0.24\linewidth, height=0.18\linewidth]{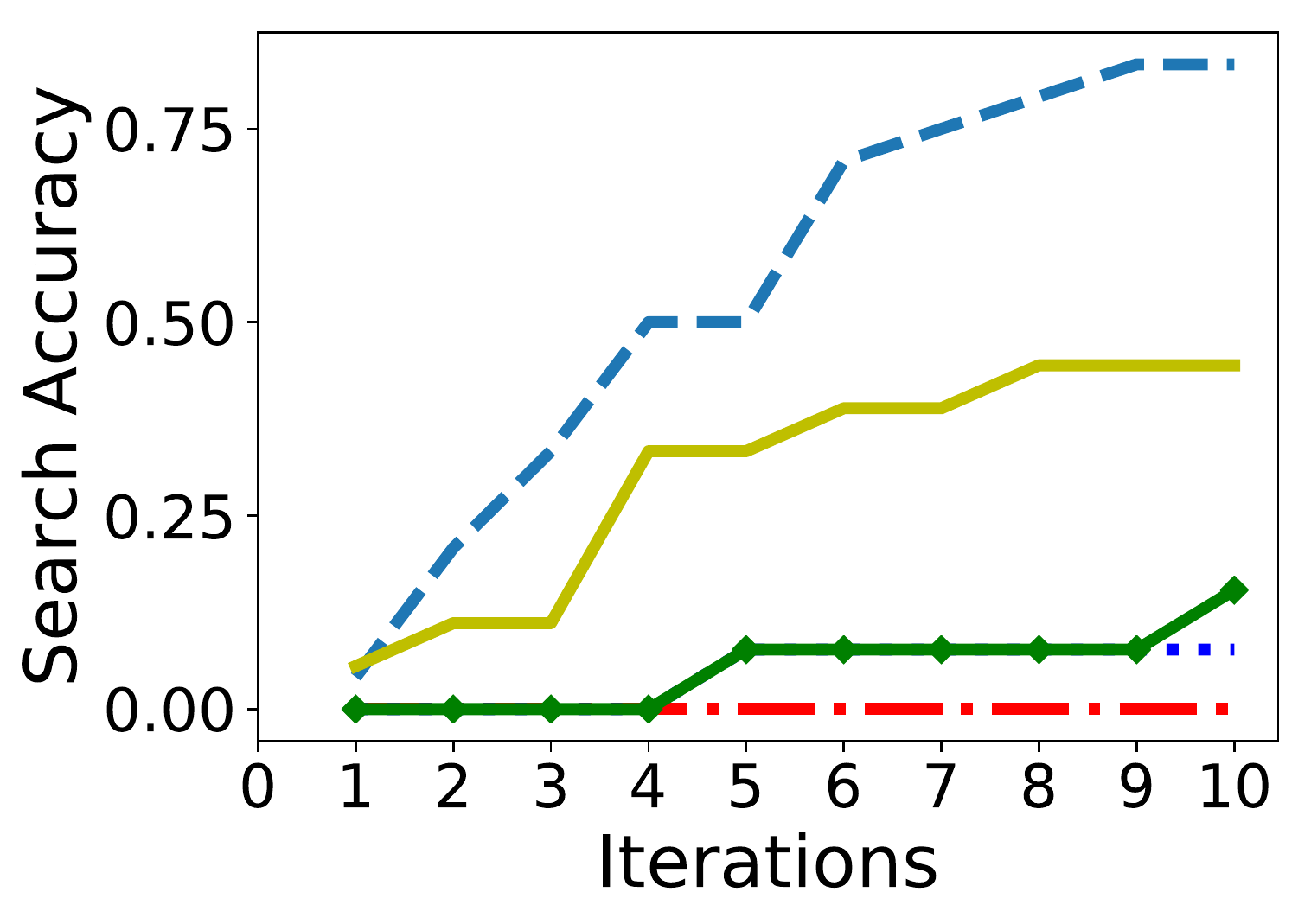}
   \hspace{0.01925\linewidth}
   \includegraphics[width=0.24\linewidth, height=0.18\linewidth]{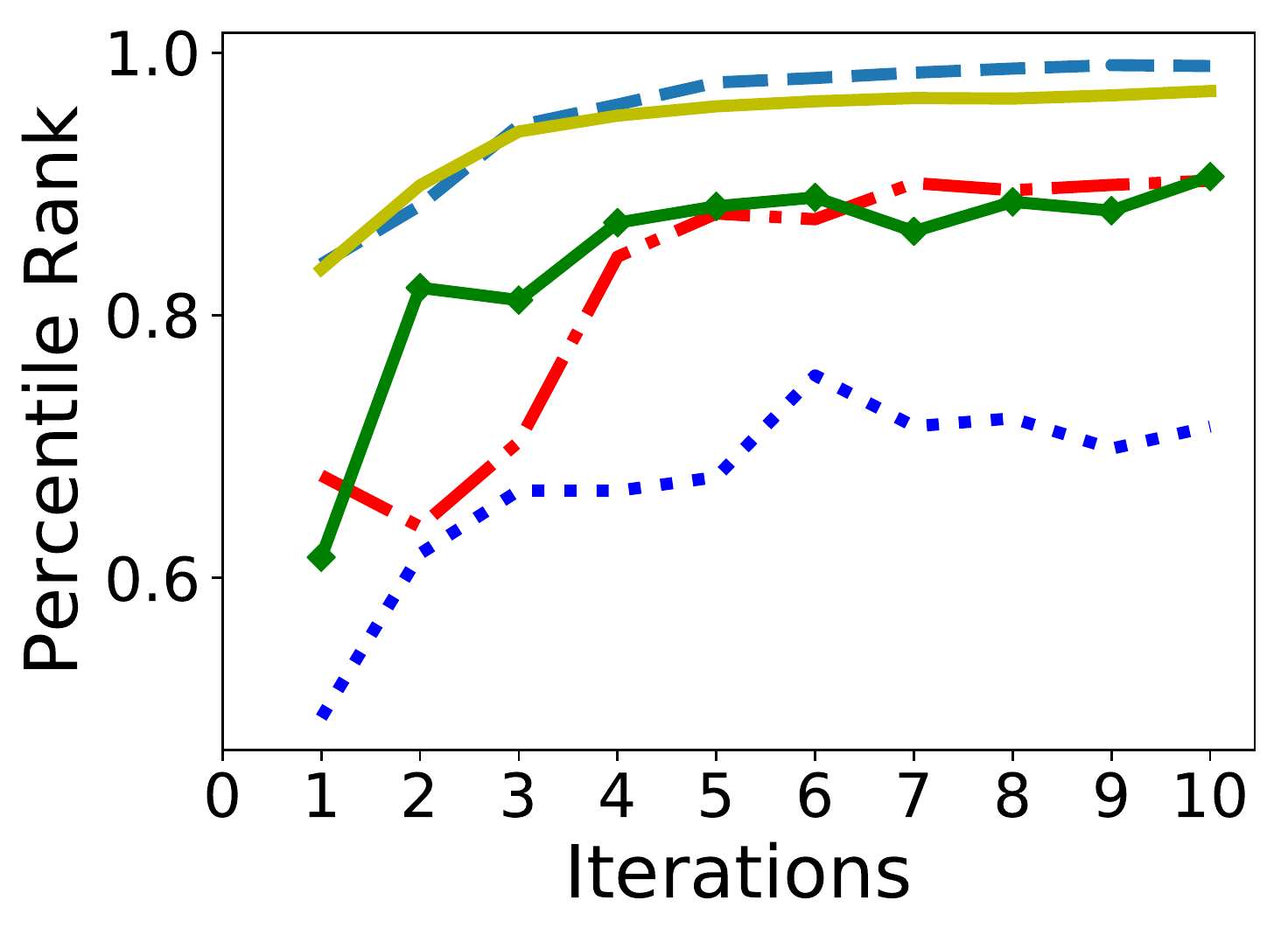}
   \includegraphics[width=0.24\linewidth, height=0.18\linewidth]{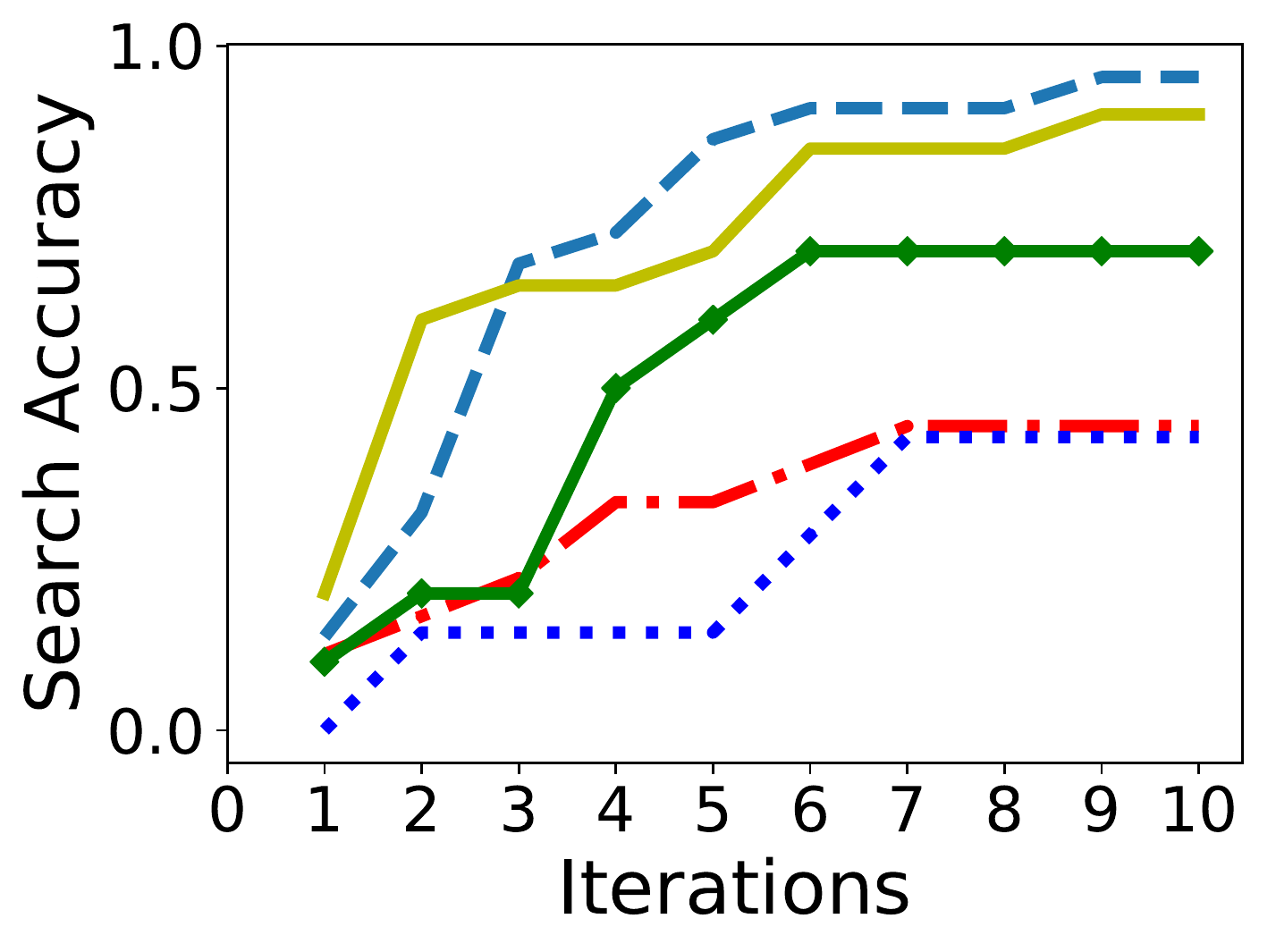}
    \caption{\small  \label{fig:wh_plot} Texture Interactive Search (TIS) Percentile Rank and Search Accuracy results on \emph{ElBa} (first two plots) and E-DTD (last two plots). On the x axis the number of feedback iterations. On the y axis the Percentile Rank index/Search accuracy score.}
\end{figure}

In this experiment we follow the Whittlesearch (WH) feedback scheme~\cite{kovashka2012whittlesearch,kovashka2015whittlesearch} to search a texture among a large repository. It can be considered a coarse-to-fine user-initiated and iterative search, with each iteration at time $t=1,\ldots,T$ presenting on a GUI the \emph{target} image, simulating the user's envisioned picture, and a \emph{reference set} $\mathcal{T}_t$ of $n=8$ images the user has to interact with by giving a feedback. At each iteration, top-ranked images are shown, until the target is ranked in the top $n$ images, or the maximum iteration $T=10$ is completed. 
The WH scheme enriches traditional binary relevance feedback mechanism~\cite{kovashka2015whittlesearch} by allowing the user to whittle away specific irrelevant portions of the visual feature space, pinpointing \emph{how} different one image in $\mathcal{T}_t$ is w.r.t. the target by using relative comparisons ("more", "equally", "less"), on a provided set of attributes. 
To prove that introducing Texel-Att attributes is beneficial to better describe textures, we set up a task of Interactive Image Search following the non-Active WhittleSearch variant~\cite{kovashka2015whittlesearch}, using our 36 attributes (see Table~\ref{tab:descriptorSize}) estimated on top of the \textit{detection} step. We compare with the attributes  extracted from \textit{ground-truth} annotations to understand how much a more accurate estimation leads to a better performance. We also compare with the 47 DTD attributes~\cite{cimpoi2014describing} employed here in their \emph{relative} form (i.e., each attribute has a ranking function indicating how much it is expressed in the image)~\cite{parikh2011relative} and the 6 Tamura~\cite{tamura1978textural} attributes (\emph{coarseness, contrast, directionality, linelikeness, regularity, roughness}). In particular we compare with three different variants: the 47 DTD attributes, the 6 Tamura attributes and the 47+6 DTD+Tamura attributes. The last combination is the most appealing, since the DTD attributes indicate the content of a texture (\emph{i.e.}, dots) and the Tamura attributes models low-level characteristics similar in spirit to ours (\emph{e.g.}, regularity) but computed on the pixels and not on texels.
For all of these, each user is presented with a randomly chosen target image from the database
. The goal is to navigate the database until the target image is found (i.e. it becomes one of the top $n$ most relevant images in the database). A total of 50 unacquainted users participated in the study (mean age: 24, std: 1), after having performed a brief individual training session on the use of the interface. Each user had three trials and performances are averaged. Users were partitioned equally among the five approaches taken into consideration in this experiment. 
Following~\cite{kovashka2015whittlesearch}, performance is measured using the percentile rank of the target image (i.e. the fraction of the database images ranked below the target) after a fixed number of interaction steps. The closer to 100\%, the better the result. 
We also compute Search Accuracy: by considering 40 images as the size of a typical image search page~\cite{kovashka2015whittlesearch}, a texture is considered as "found" by the search if it is ranked among the first 40 images. Keeping this in mind we find that on the E-DTD dataset we are able to individuate within 10 iterations the desired texture in the 90\% of cases while using the most performing variant's (DTD+Tamura) attributes accuracy drops down to 71\%. On the \emph{ElBa} dataset, which is much more challenging on this task due to the larger size of the pool of images and the finer grained nature of the textures, we are able to reach 44\% accuracy, while the DTD+Tamura attributes reach only 15\%.
The plot in Fig.~\ref{fig:wh_plot} shows that Texel-Att has the best performance at any iteration. In addition, a good performance is preserved even in the case of predicted attributes, confirming the robustness of the approach against imperfect texel detection. Finally, on average we are able to individuate the desired texture more often with our approach than with other techniques.

\section{Discussion}
This paper presents a new way to describe textures, adopting attributes which focus on texels. The proposed framework, Texel-Att, can successfully describe and classify patterns that are not well-handled by the existing texture analysis approach, as demonstrated by the experimental results, supported also by two new datasets, ElBa and E-DTD. In addition, Texel-Att is showed to be highly effective for image search, paving the way to fashion and graphic design applications.
The current implementation has much room for improvement, being trained with few texel types (circles, lines, polygons) and 2D patterns with limited distortions. 
In fact, the modular design of the framework makes it easy to customize it for handling different kinds of element-based texture, as it is just a matter of changing the detector to localize and group different texel types, and changing the invariance properties of the layout attributes to handle larger distortions. 
\\
\\
\textbf{Acknowledgements:} This work has been partially supported by the project of the Italian Ministry of Education, 
Universities and Research (MIUR) "Dipartimenti di Eccellenza 2018-2022", and has been partially supported
by the POR FESR 2014-2020 Work Program (Action 1.1.4,
project No.10066183). We also thank Nicolò Lanza for assistance with Substance Designer software.

\bibliography{main}
\end{document}


\title{\emph{Texel-Att}: Representing and Classifying Element-based Textures by Attributes - Supplementary Material}

\addauthor{Marco Godi*}{marco.godi@univr.it}{1}
\addauthor{Christian Joppi*}{christian.joppi@univr.it}{1}
\addauthor{Andrea Giachetti}{andrea.giachetti@univr.it}{1}
\addauthor{Fabio Pellacini}{pellacini@di.uniroma1.it}{2}
\addauthor{Marco Cristani}{marco.cristani@univr.it}{1}

\addinstitution{
 Department of Computer Science\\
 University of Verona\\
 Strada le Grazie 15, Italy
}
\addinstitution{
 Sapienza, University of Rome\\
Rome,\\
 Italy
 {\textcolor{white}{.}} \\
{\textcolor{white}{.}} \\
{\textcolor{white}{.}} \\
{\textcolor{white}{.}} \\
{\footnotesize * These authors contributed equally to this work.} \\
}

\runninghead{Godi, Joppi}{Texel-Att: Representing [...] element-based textures}

\def\eg{\emph{e.g}\bmvaOneDot}
\def\Eg{\emph{E.g}\bmvaOneDot}
\def\etal{\emph{et al}\bmvaOneDot}


\maketitle
\vspace{-1em}
\section{Organization of the Supplementary Material}
The supplementary material is organized as follows: qualitative results of the texels' detection on the E-DTD and the \emph{ElBa} datasets are shown in Sec.~\ref{sec:det}. In Sec.~\ref{sec:rank} we show, for each of the attributes we selected, the images corresponding to the minimum, the median and the maximum intensity of that attribute, both on the \emph{ElBa} dataset and within some E-DTD classes. In Sec.~\ref{sec:relatt} we show the ranking accuracies in the \emph{ElBa} and E-DTD dataset for all attributes, for our approach and the comparative one. Finally, we clarify the theoretical foundations for some attributes in Sec.~\ref{sec:symmetry}

\section{Detection Results}\label{sec:det}

\subsection{E-DTD dataset}
In this section, we show an excerpt of the 12 E-DTD classes where detection results (bounding boxes) are shown in green for correct detections, in red for false positives, in blue for false negatives.

\begin{figure}[htb]
\centering
\includegraphics[width=\textwidth]{figures_sup_material/collage-edtd-det.pdf}
\caption{\small First column: Banded; Second column: Chequered; Third Column: Dotted.}
\end{figure}

\begin{figure}[htb]
\centering
\includegraphics[width=\textwidth]{figures_sup_material/collage-edtd-det2.pdf}
\caption{\small First column: Grid; Second column: Honeycombed; Third Column: Lined.}
\end{figure}

\begin{figure}[htb]
\centering
\includegraphics[width=\textwidth]{figures_sup_material/collage-edtd-det3.pdf}
\caption{\small First column: Meshed; Second column: Perforated; Third Column: Polka-Dotted.}
\end{figure}

\begin{figure}[htb]
\centering
\includegraphics[width=\textwidth]{figures_sup_material/collage-edtd-det4.pdf}
\caption{\small First column: Studded; Second column: Waffled; Third Column: Woven.}
\end{figure}

\clearpage
\subsection{\emph{ElBa} dataset}
In this section, we show an excerpt of the \emph{ElBa} dataset where detection results (bounding boxes) are shown in green for correct detections, in red for false positives, in blue for false negatives.
\begin{figure}[h!]
\centering
\includegraphics[width=\textwidth]{figures_sup_material/collage-elba-det.pdf}
\caption{\small Detection results on \emph{ElBa} dataset.}
\end{figure}

\clearpage
\section{Attributes Ranking}\label{sec:rank}

\subsection{\emph{ElBa} dataset}
In this section, we show an excerpt of the \emph{ElBa} dataset where images are sorted according to a specific attribute.

\begin{figure}[htb]
\centering
\includegraphics[width=\textwidth]{figures_sup_material/area.pdf}
\caption{\small From texture with the smallest average texels' area (left) to the texture  with the biggest one (right).}
\vspace{-1.5em}
\end{figure}

\begin{figure}[htb]
\centering
\includegraphics[width=\textwidth]{figures_sup_material/orientation.pdf}
\caption{\small From texture with the vertical orientation (left) to the texture with the horizontal one (right).}
\vspace{-1.5em}
\end{figure}

\begin{figure}[htb]
\centering
\includegraphics[width=\textwidth]{figures_sup_material/homog.pdf}
\caption{\small From texture with the highest elements' homogeneity (left) to the texture with the lowest one (right).}
\vspace{-1.5em}
\end{figure}

\begin{figure}[htb]
\centering
\includegraphics[width=\textwidth]{figures_sup_material/r0.pdf}
\caption{\small From texture with the most regular layout disposition (left) to the texture with the least regular one (right).}
\vspace{-1.5em}
\end{figure}

\newpage
\subsection{E-DTD dataset}
In this section, we show an excerpt of the E-DTD dataset where images from a specific class are sorted according to a specific attribute.
\subsubsection{Banded Class}
\begin{figure}[htb]
\centering
\includegraphics[width=\textwidth]{figures_sup_material/area_banded.pdf}
\caption{\small From texture with the smallest average texels' area (left) to the texture  with the biggest one (right).}
\vspace{-1.5em}
\end{figure}

\begin{figure}[htb]
\centering
\includegraphics[width=\textwidth]{figures_sup_material/orientation_banded.pdf}
\caption{\small From texture with the vertical orientation (left) to the texture with the horizontal one (right).}
\vspace{-1.5em}
\end{figure}

\begin{figure}[htb]
\centering
\includegraphics[width=\textwidth]{figures_sup_material/r0_banded.pdf}
\caption{\small From texture with the most regular layout disposition (left) to the texture with the least regular one (right).}
\vspace{-1.5em}
\end{figure}

\newpage
\subsubsection{Chequered Class}
\begin{figure}[htb]
\centering
\includegraphics[width=\textwidth]{figures_sup_material/area_cheq.pdf}
\caption{\small From texture with the smallest average texels' area (left) to the texture  with the biggest one (right).}
\vspace{-1.5em}
\end{figure}

\begin{figure}[htb]
\centering
\includegraphics[width=\textwidth]{figures_sup_material/orientation_cheq.pdf}
\caption{\small From texture with the vertical orientation (left) to the texture with the horizontal one (right).}
\vspace{-1.5em}
\end{figure}

\begin{figure}[htb]
\centering
\includegraphics[width=\textwidth]{figures_sup_material/r0_cheq.pdf}
\caption{\small From texture with the most regular layout disposition (left) to the texture with the least regular one (right).}
\vspace{-1.5em}
\end{figure}

\newpage
\subsubsection{Dotted Class}
\begin{figure}[htb]
\centering
\includegraphics[width=\textwidth]{figures_sup_material/area_dotted.pdf}
\caption{\small From texture with the smallest average texels' area (left) to the texture  with the biggest one (right).}
\vspace{-1.5em}
\end{figure}

\begin{figure}[htb]
\centering
\includegraphics[width=\textwidth]{figures_sup_material/orientation_dotted.pdf}
\caption{\small From texture with the vertical orientation (left) to the texture with the horizontal one (right).}
\vspace{-1.5em}
\end{figure}

\begin{figure}[htb]
\centering
\includegraphics[width=\textwidth]{figures_sup_material/r0_dotted.pdf}
\caption{\small From texture with the most regular layout disposition (left) to the texture with the least regular one (right).}
\vspace{-1.5em}
\end{figure}

\newpage
\section{Relative Attributes Experiment Results}\label{sec:relatt}
In this section, we show ranking accuracies for each attribute both for the E-DTD dataset and the \emph{ElBa} dataset.

\begin{table}[htb]

\scalebox{0.75}{
\begin{tabular}{|l|p{40mm}|l|l|}
\hline
Attribute Name & Attribute Interpretation &  FV-CNN Ranking SVM & Texel-Att \\
 & & Accuracy & Accuracy \\
\hline
\hline
\textit{ \% of Circle Texels} 	& Portion of texels of circular shape	&  0.9040 & \textbf{ 1.00}\\\hline
\textit{ \% of Line Texels} 	& Portion of texels of linear  shape	&  0.8966 & \textbf{ 1.00}\\\hline
\textit{ \% of Polygon Texels} 	& Portion of texels of polygonal shape 	&  0.8661 & \textbf{ 1.00}\\\hline
\textit{ Components of Background Color Histogram} 	& Portion of the background colored in a particular color &  0.6674 & \textbf{ 0.9071}\\\hline
\textit{ Texel Area} 	& Average texel size	&  0.6552 & \textbf{ 0.8571}\\\hline
\textit{ Texel Density} 	& Texel density		&  0.9055 & \textbf{ 0.9514}\\\hline
\textit{ Components of Texel Orientation Histogram} 	& Portion of texels oriented on a particular direction 	&  0.6632 & \textbf{ 0.6916}\\\hline
\textit{ Components of Texel Color Histogram} 	& Portion of texels colored in a particular color	&  0.6674 & \textbf{ 0.9466}\\\hline
\textit{ Texel Homogeneity} 	& Degree of uniformity in the density of texels 	&  0.7991 & \textbf{ 0.9029}\\\hline
\textit{ Local Symmetry} 	& Degree of symmetry in between nearby texels	&  0.5472 & \textbf{ 0.6429}\\\hline
\textit{ Traslational Symmetry} 	& Similarity of positioning between different pairs of texels	&  0.5082 & \textbf{ 0.5957}\\\hline
\textit{ Components of Traslation Histogram} 	& Portion of texel pairs positioned on a particular direction &  0.5747 & \textbf{ 0.6881}\\\hline
\end{tabular}
}
\vspace{0.8em}
\caption{Ranking accuracy for every attribute for the E-DTD dataset. Our approach performs better on every attribute. Histograms are averaged over each single component for the sake of clarity.}

\end{table}

\begin{table}[]

\scalebox{0.75}{
\begin{tabular}{|l|p{40mm}|l|l|}
\hline
Attribute Name & Attribute Interpretation &  FV-CNN Ranking SVM & Texel-Att \\
 & & Accuracy & Accuracy \\
\hline
\hline
\textit{ \% of Circle Texels} 	& Portion of texels of circular shape	&  0.7271 & \textbf{ 0.9886}\\\hline
\textit{ \% of Line Texels} 	& Portion of texels of linear  shape	&  0.8156 & \textbf{ 0.9843}\\\hline
\textit{ \% of Polygon Texels} 	& Portion of texels of polygonal shape 	&  0.7166 & \textbf{ 0.9814}\\\hline
\textit{ Components of Background Color Histogram} 	& Portion of the background colored in a particular color &  0.6302 & \textbf{ 0.9308}\\\hline
\textit{ Texel Area} 	& Average texel size	&  0.7016 & \textbf{ 0.9386}\\\hline
\textit{ Texel Density} 	& Texel density		&  0.8186 & \textbf{ 0.9686}\\\hline
\textit{ Components of Texel Orientation Histogram} 	& Portion of texels oriented on a particular direction 	&  0.6347 & \textbf{ 0.8367}\\\hline
\textit{ Components of Texel Color Histogram} 	& Portion of texels colored in a particular color	&  0.6291 & \textbf{ 0.9949}\\\hline
\textit{ Texel Homogeneity} 	& Degree of uniformity in the density of texels 	&  0.7706 & \textbf{ 0.9686}\\\hline
\textit{ Local Symmetry} 	& Degree of symmetry in between nearby texels	&  0.7106 & \textbf{ 0.8171}\\\hline
\textit{ Traslational Symmetry} 	& Similarity of positioning between different pairs of texels	&  0.6537 & \textbf{ 0.8714}\\\hline
\textit{ Components of Traslation Histogram} 	& Portion of texel pairs positioned on a particular direction &  0.6895 & \textbf{ 0.7395}\\\hline
\end{tabular}
}
\vspace{0.8em}
\caption{\small Ranking accuracy for every attribute for the \emph{ElBa} dataset. Our approach performs better on every attribute. Histograms are averaged over each single component for the sake of clarity.}
\end{table}

\newpage
\section{Mathematical description of symmetry attributes} \label{sec:symmetry}
In this section we provide a more detailed explanation for the symmetry attributes.

Given the spatial distribution of the centroids of the detected dots and polygons, we characterize their average center-reflective and translational symmetry as follows. 
For the average reflective symmetry, we consider for each centroid a 4-points neighborhood, reflect the points of the neighborhood with respect to the center and estimate the distance from the closest centroid grid point.
The average symmetry score is
\[
S(R) = \frac{1}{N} \sum_{i=1}^{N} \sum_{j=1}^{4} \left | \vec{R_i}( \vec{n}_{ij}) - \vec{c}_{i} \right |
\]
Where N is the total number of centroids in the group, $ \vec{c}_i$ is a centroid, $R_i$ is the reflection around centroid $i$, $\vec{n}_{ij}$ is the $j-th$ nearest neighbor of the centroid $i$ (Fig. \ref{fig:symm1} left).

For the average translational symmetry, we consider for each centroid a 4-points neighborhood, translate the points of the neighborhood of all the $\vec{t_j}$ vectors joining it with the neighbors and estimate the distance from the closest centroid grid point.
The average symmetry score is
\[
S(T) = \frac{1}{4N} \sum_{i=1}^{N} \sum_{j=1}^{4} \sum_{k=1}^{4} \left |  \vec{n}_{ik} - \vec{c}_{i} +\vec{t}_{j} \right |
\]
Where N is the total number of centroids in the group, $ \vec{c}_i$ is a centroid, $R_i$ is the reflection around centroid $i$, $\vec{n}_{ij}$ is the $j-th$ nearest neighbor of the centroid $i$ (Fig. \ref{fig:symm1} right).


\begin{figure}
\centering
 \includegraphics[width=0.4\linewidth]{figures_sup_material/symm1.eps}   \includegraphics[width=0.4\linewidth]{figures_sup_material/symm2.eps}
    \caption{
     \label{fig:symm1} Symmetry scores are evaluated as an average of local self-similarity of elements' centroid patterns after translation of 4-point neighborhoods of point pairs vectors included in the neighborhood (left) and after reflection with respect to the central point (right).}
\end{figure}